\documentclass[letterpaper, 10 pt, conference]{styles/ieeeconf}
\IEEEoverridecommandlockouts
\usepackage{algorithm,algorithmic}

\usepackage[dvips]{graphicx}
\usepackage{graphics}
\graphicspath{{./figures/}}
\usepackage{multirow}
\usepackage{textcomp}
\usepackage{subcaption}
\usepackage{comment}
\usepackage{cite}
\usepackage{url}
\usepackage{wrapfig}
\usepackage{color, xcolor}
\usepackage{flushend}
\usepackage{verbatim}
\usepackage{amsfonts}

\newcommand{\dofs}{{\sc dof}s}
\newcommand{\hasap}{HASP}
\newcommand{\prm}{Basic PRM}

\newcommand{\maprm}{MAPRM}

\newcommand{\rrt}{RRT}

\newcommand{\drbrrt}{Dynamic Region Sampling with RRT}
\newcommand{\drprm}{Dynamic Region Sampling with PRM}
\newcommand{\lazyprm}{Lazy PRM}
\newcommand{\customlazyprm}{Partial Lazy PRM}
\newcommand{\cspace}{\mathcal{C}_{space}}
\newcommand{\cfree}{\mathcal{C}_{free}}
\newcommand{\cobst}{\mathcal{C}_{obst}}

\title{\LARGE \bf
Hierarchical Planning with Annotated Skeleton Guidance}

\author{Diane Uwacu$^{1}$, Ananya Yammanuru$^{2}$, Marco Morales$^{2,3}$, Nancy M. Amato$^{2}$%
\thanks{$^{1}$Diane Uwacu is with the Department of Computer Science and Engineering at Texas A\&M University, College Station, TX, USA
        {\tt\small duwacu@tamu.edu}}%
\thanks{$^{2}$ Ananya Yammanuru, Marco Morales and Nancy M. Amato are with the Department of Computer Science at the University of Illinois at Urbana-Champaign, Urbana, IL, USA
        {\tt\small (ananyay2, moralesa, namato)@illinois.edu}}%
\thanks{$^{3}$ Marco Morales is also with the Department of Computer Science at Instituto Tecnológico Autónomo de México (ITAM), Mexico City, México}%
}

\begin{document}

\maketitle
\thispagestyle{plain}
\pagestyle{plain}

\begin{abstract}
We present a hierarchical skeleton-guided motion planning algorithm to guide mobile robots. 
A good skeleton maps the connectivity of the subspace of c-space containing significant degrees of freedom and is able to guide the planner to find the desired solutions fast.
However, sometimes the skeleton does not closely represent the free c-space, which often misleads current skeleton-guided planners.
The hierarchical skeleton-guided planning strategy gradually relaxes its reliance on the workspace skeleton as $\cspace$ is sampled, thereby incrementally returning a sub-optimal path, a feature that is not guaranteed in the standard skeleton-guided algorithm.
Experimental comparisons to the standard
skeleton guided planners and other lazy planning strategies show significant improvement in roadmap construction run time while maintaining path quality for multi-query problems in cluttered environments. 

\end{abstract}

\section{Introduction}

Motion Planning algorithms are applied to a multitude of applications ranging from robotic navigation \cite{englot2012AAAI} to microbiology simulations \cite{baghvsl-urfpdl}. 
The sampling-based probabilistic roadmap method (\prm) \cite{kslo-prpp-96} introduced in the late nineties efficiently deals with the intractability of motion planning problems by approximating high-dimensional spaces. It led to several innovations to improve its performance \cite{abdjv-obprm-98, hk-fuwmapp-00}.
The two main targets for improvement are 1) efficiency in finding a sub-optimal solution fast, and 2) coverage that leads to reusability for multi-query problems.

We recently introduced \drbrrt~\cite{dsba-drbrrt-16} and \drprm \cite{suda-tgrcdrs-20}, two workspace-guided motion planning strategies, as a reliable way to guide planners, especially in applications where the workspace is closely tied to the planning space. The workspace skeleton graph provides a minimal representation of the connectivity of the free space, on top of which the planning algorithm defines dynamic sampling regions that are guided by the edges of the skeleton. However, both strategies rely heavily on the workspace skeleton by constraining roadmap expansion to the intermediate points of the skeleton edges. These strategies also often waste time exploring regions that the skeleton has already validated.

In this work, we introduce the Hierarchical Annotated Skeleton Planning (\hasap) strategy for finding workspace-guided paths in complex environments. 
Similar to \cite{suda-tgrcdrs-20}, \hasap~builds a skeleton-guided roadmap. But, in contrast to the previous work, \hasap~uses the skeleton to find paths hierarchically. It initially searches for paths that are indicated by the skeleton and easy-to-find, and progressively searches for more difficult paths (those that require more planning) according to user-defined preferences. 
Additionally, the \hasap~method delays validation, similar to \cite{bk-ppulp-00}, to reduce the number of collision detection calls that are required to build a decent roadmap.\par

Specifically, our contribution is a new skeleton-guided strategy that hierarchically focuses planning effort according to user-defined criteria.
The algorithm improves its reliance on the workspace skeleton as the environment is explored. Additionally, thanks to skeleton annotations, the strategy can be adapted to fit different applications of motion planning.
Our experimental analysis shows how the \hasap~method achieves improved run time, efficiency, and scalability compared to related sampling-based planning strategies.

\section{Preliminaries and Previous Work}
\label{sec:previousWork}
\begin{figure*}[!t]
  \centering
\begin{subfigure}{.24\textwidth}
  \centering
  \includegraphics[width=1\textwidth]{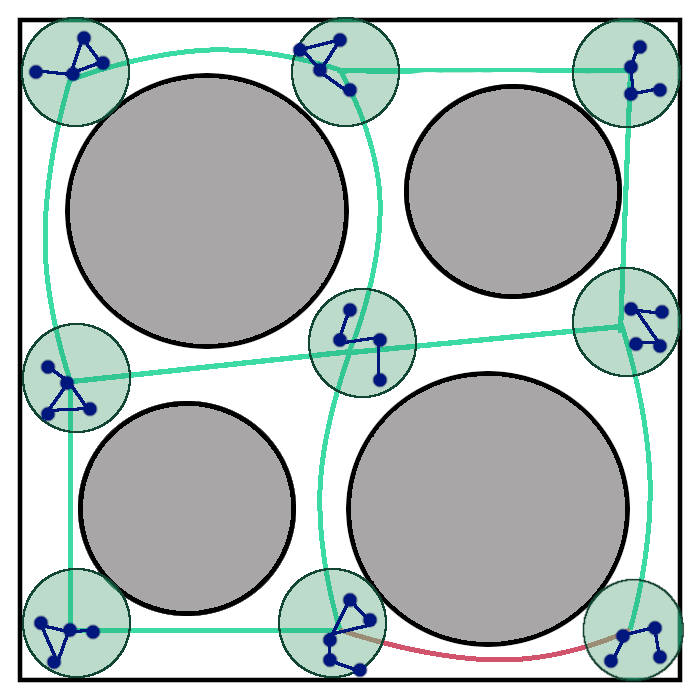}
  \caption{}
  \label{fig:drprm_init}
\end{subfigure}
  \centering
\begin{subfigure}{.24\textwidth}
  \centering
  \includegraphics[width=1\textwidth]{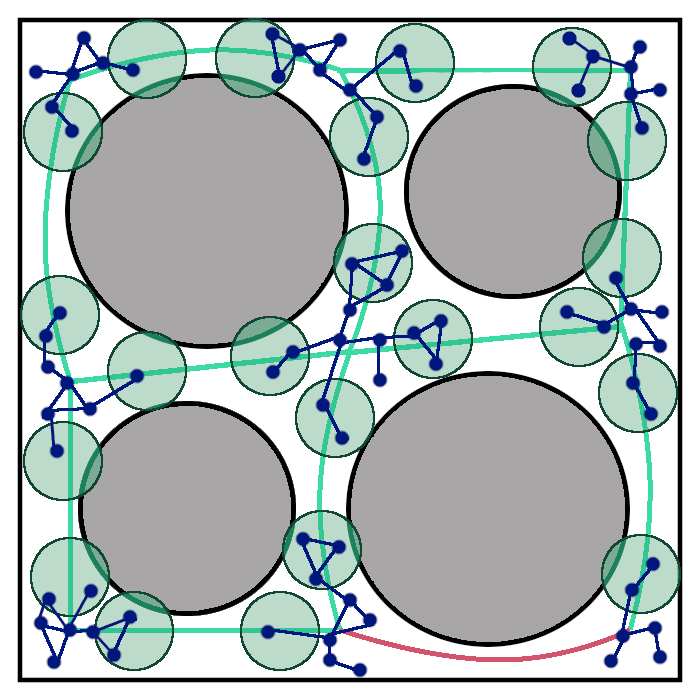}
  \caption{}
  \label{fig:drprm_expand}
\end{subfigure}
\centering
\begin{subfigure}{.24\textwidth}
  \centering
  \includegraphics[width=1\textwidth]{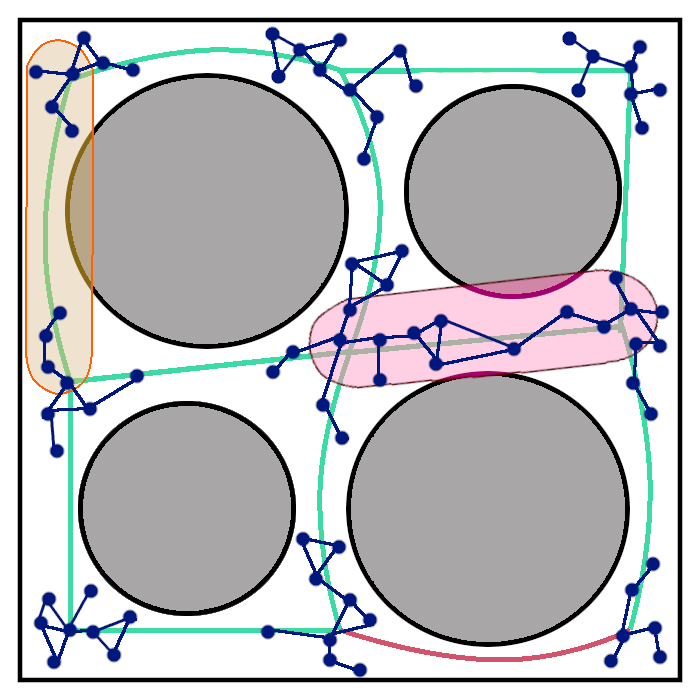}
  \caption{}
  \label{fig:drprm_bridge}
\end{subfigure}
\centering
\begin{subfigure}{.24\textwidth}
  \centering
  \includegraphics[width=1\textwidth]{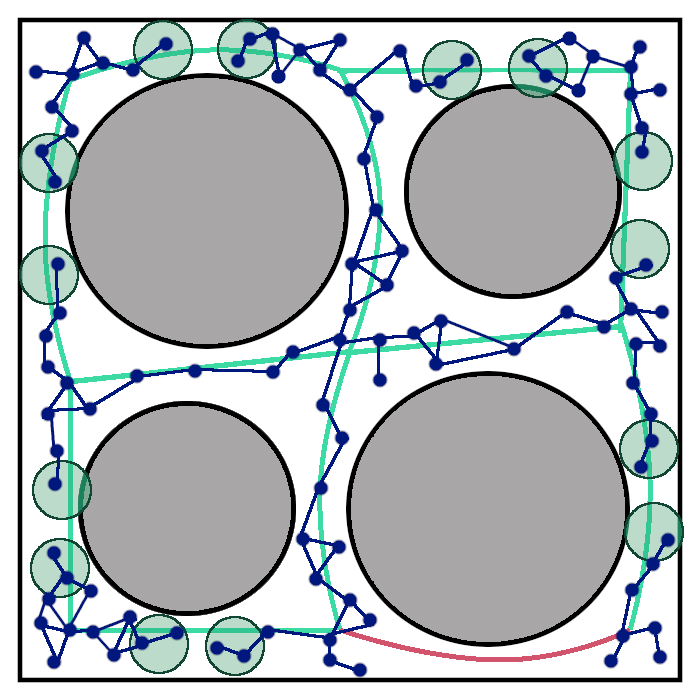}
  \caption{}
  \label{fig:drprm_fullrdmp}
\end{subfigure}
\caption{An illustration of Dynamic Region sampling with PRM. Obstacles are shown in gray. The workspace skeleton is shown in purple. (a) The algorithm
samples initial connected components (blue) in regions (green) around each skeleton vertex. (b) Sampling regions expand outward along the skeleton edges. We
depict the regions in the location where samples were generated for clarity; in the actual algorithm the regions advance past the newly generated samples. (c) An
illustration of two edge segments. The red-shaded segment has a single local component which is also a bridge. The orange-shaded segment has two distinct local
components. (d) The components in the middle tunnels successfully connect to form bridges, and their regions are released. The outer passages are still expanding.}
 \label{fig:drprm}
\end{figure*}

The pose of all the components of a robot can be fully determined by its configuration, and the set of all
  configurations within an environment is the \textit{configuration space}, or $\cspace$. The dimensionality of
  $\cspace$ is equivalent to the robot's degrees of freedom (\dofs).
$\cspace$ is often partitioned
into free space ($\cfree$) and obstacle space ($\cobst$).

Given a start $q_s$ and a goal $q_g$ configurations or regions, that define a query, the
motion planning problem consists of finding a continuous
path $\tau$ in $\cfree$ between the start $q_s$ and the goal $q_g$ as follows:
  \begin{equation}
    \tau : \mathbb{R} \rightarrow \cfree
    \mid
    \ \tau(0) = q_s
    ,\ \tau(1) = q_g
  \end{equation}

Generally, a representation of $\cobst$ cannot be computed. The validity of a single configuration $q$ can be determined with the help of collision detection methods. This allows sampling-based motion planning strategies to randomly explore $\cspace$ without having to compute $\cobst$.
  
\subsection{Sampling-based Algorithms}

Sampling based algorithms approximate $\cspace$ by generating random robot configurations and connecting them to build a roadmap graph (like the probabilistic roadmap (\prm)
\cite{kslo-prpp-96}) or a tree (like the rapidly-exploring random
trees (\rrt) \cite{l-rrtntpp-1998}). In this work we focus on \prm~and its variants whose roadmaps can be searched to find valid
paths for multiple queries. 

Because it explores an environment randomly, the \prm~algorithm suffers from the narrow passage problem in environments with low expansiveness \cite{hlk-pfprp-06}. This problem has been typically mitigated by biasing sampling. For example,
obstacle-based algorithms \cite{abdjv-obprm-98, bsa-lbobprm-01, ytea-uudobp-12} sample close to obstacle surfaces, and medial axis based algorithms
\cite{lta-gfsmafs-03, yb-asdppbama-04, hk-fuwmapp-00} sample along the medial axis of the environment and find paths with maximized obstacle
clearance. Unfortunately, these strategies are often costly because they require more collision detection calls to push configurations to the desired positions.

\subsection{Guided Motion Planning}
To maximize the chance of sampling free $\cspace$, guided planning has gained traction. A User-Guided Planning Strategy \cite{dsja-arbsfcrc-14} allows the user to define and manipulate workspace sampling regions that the planner follows in real time to explore the planning space inside the user-defined region. The planner relies on the user’s intuition to plan inside narrow passages and find intuitive paths faster. The success of this approach inspired solutions that use topological workspace guidance.

Workspace guidance involves using the workspace structure to direct $\cspace$ exploration. 
Some approaches \cite{bo-uwignsprp-04, kh-wisprp-04} decompose the workspace to specifically target narrow
passages. They partition the workspace into
cells that can be used to bias the sampling process.
Unfortunately, these methods generally suffer from oversampling regions because the cells are predetermined.

Our group has used workspace skeletons to guide RRT \cite{dsba-drbrrt-16} and PRM \cite{suda-tgrcdrs-20} algorithms. These strategies guide the planner to sample inside dynamic regions created along the workspace skeleton's edges. The \drbrrt~ strategy creates a rapidly-exploring random tree along the skeleton, while the \drprm~method builds a probabilistic roadmap\cite{suda-tgrcdrs-20} using the skeleton.
The experimental results from both methods show that skeleton-guided planners are more efficient than their non-guided counterparts, but their performance depends on the quality of the workspace skeleton. By guiding the sampling regions along the skeleton edges, these algorithms can only find a solution in reasonable time if the solution exists in workspace and is mapped by the skeleton.
In addition, these planners are only guided by the workspace connectivity indicated by the skeleton, and do not fully utilize properties specific to the workspace that could be relevant to the exploration of $\cspace$. In contrast, in this work we propose to use the information on skeleton edges to limit the level of local exploration done in the connecting regions.

\drprm \cite{suda-tgrcdrs-20}, the predecessor of this work (illustrated in Figure \ref{fig:drprm}) creates local components around the skeleton vertices and expand them along skeleton edges. Sampling regions are initialized at every skeleton vertex. The C-space inside those regions is explored by generating local components (Fig. \ref{fig:drprm_init}), and the regions are advanced along the intermediates of the skeleton edges as the local components expand. Regions expand incrementally, making ``small steps” from the skeleton vertex towards the center of the edge (Fig. \ref{fig:drprm_expand}). When regions begin to touch or overlap, a bridge connection is attempted, joining the two region (Fig. \ref{fig:drprm_bridge}). It is possible that a bridge cannot be made, resulting in two separate paths over the length of the skeleton edge.

\section{Method}

\begin{figure*}[ht]
  \centering
\begin{subfigure}{.24\textwidth}
  \centering
  \includegraphics[width=1\textwidth]{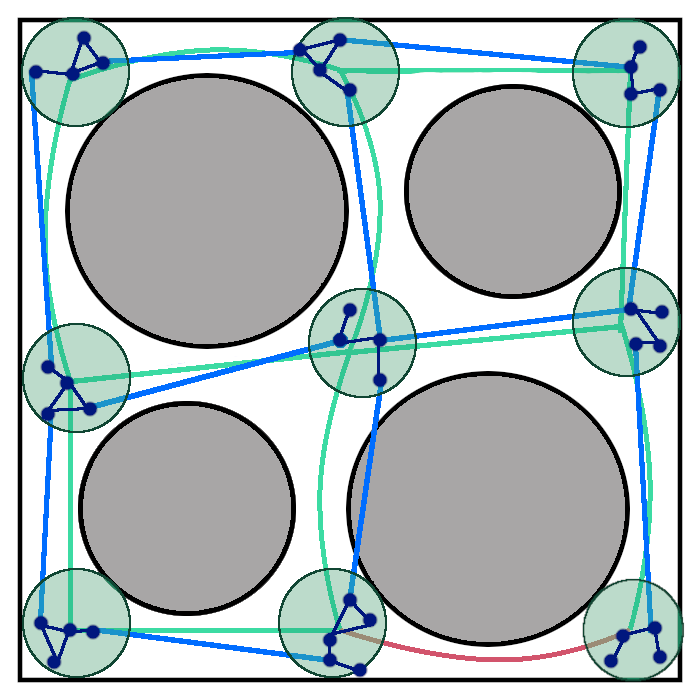}
  \caption{}
  \label{fig:lazy_draw_edges}
\end{subfigure}
  \centering
\begin{subfigure}{.24\textwidth}
  \centering
  \includegraphics[width=1\textwidth]{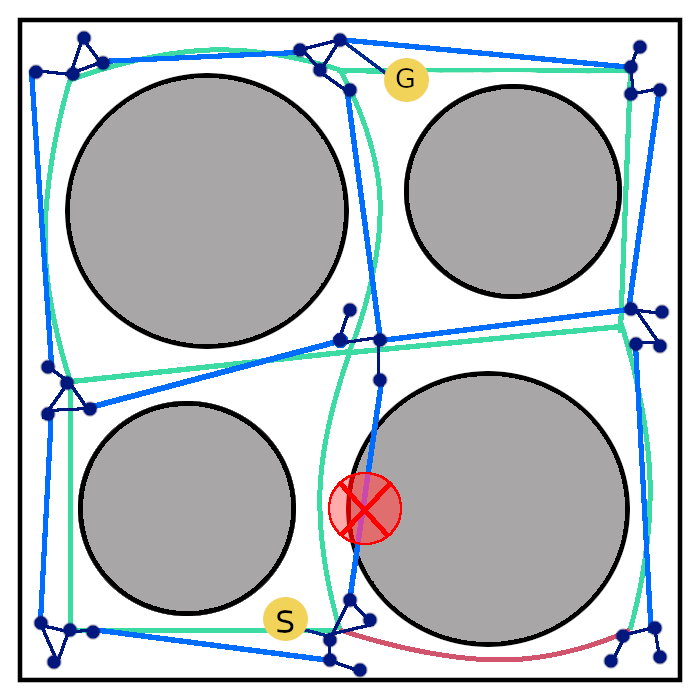}
  \caption{}
  \label{fig:delete_invalid_edge_in_path}
\end{subfigure}
\centering
\begin{subfigure}{.24\textwidth}
  \centering
  \includegraphics[width=1\textwidth]{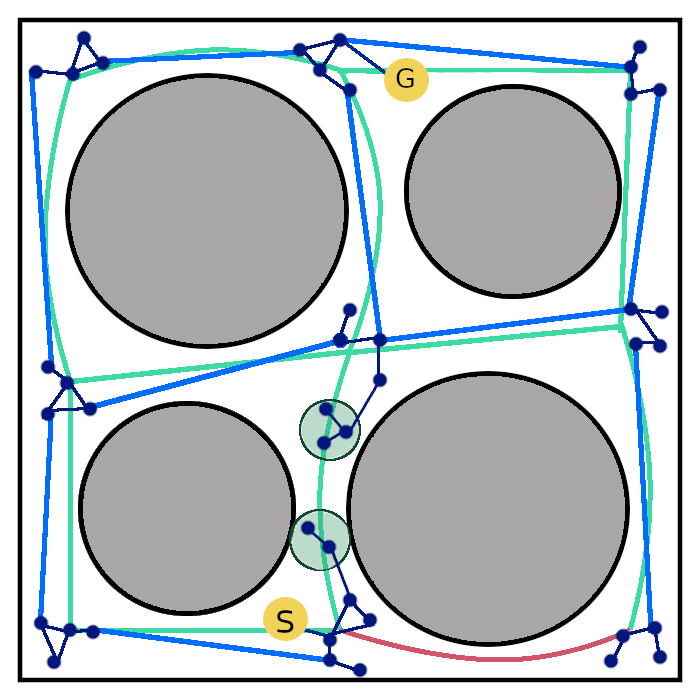}
  \caption{}
  \label{fig:expand_regions}
\end{subfigure}
\centering
\begin{subfigure}{.24\textwidth}
  \centering
  \includegraphics[width=1\textwidth]{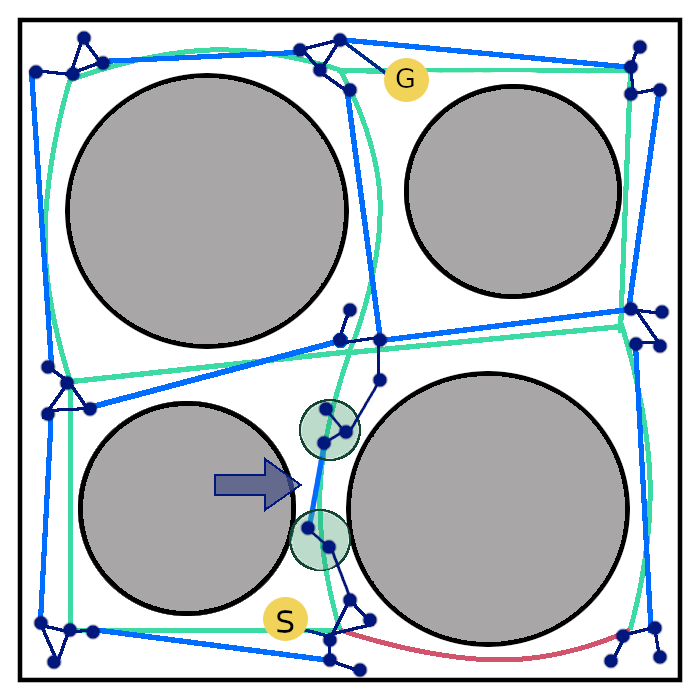}
  \caption{}
  \label{fig:lazy_draw_new_edge}
\end{subfigure}
\caption{Example execution of the \hasap~method: (a)
  Generate the clearance-annotated workspace skeleton (green edges denote sufficient clearance, red denotes insufficient clearance). Sampling regions are initialized at each skeleton vertex with enough clearance. Edges are added without validation (shown in light blue) if the corresponding skeleton edge weight meets the acceptance criteria (in this example, the skeleton edge is green).
  (b) Query the roadmap and return all valid paths. If no fully valid paths, mark conflict edges in invalid paths and return.
  (c) If a valid path was found, return it. Else, attempt to fix invalid edges by expanding sampling regions from each end of the skeleton edge corresponding to the invalid edge.
  (d) Draw an edge without validation between the expanded sampling regions. Return to step (b). }
 \label{fig:MethodExplanation}
\end{figure*}

Algorithm \ref{alg:fullalg} and Figure \ref{fig:MethodExplanation} describe our proposed \hasap~method. 
We begin by constructing a workspace skeleton, which is a unidimensional graph in the workspace such that its free space can be collapsed into the skeleton continuously \cite{article}. 
The skeleton's vertices are mapped to topological regions of the workspace and the skeleton's edges indicate the connectivity of two adjacent regions.

In Line 2, the skeleton is annotated with information that may be relevant to the $\cspace$. 
The user decides on an annotation policy to gather information from the workspace that helps guide the planner in a desired $\cspace$~exploration behavior. 
For example, in some problems, paths that maximize obstacle clearance are preferred \cite{lta-gfsmafs-03}. So, as illustrated in Fig.~\ref{fig:lazy_draw_edges}, each  skeleton vertex is annotated with a clearance value while the edges are assigned a weight corresponding to the lowest clearance value along the edge. After the skeleton is built and annotated, the planner uses it to explore the $\cspace$. This is done in three phases: building the roadmap, building the path set, and fixing the paths in the path set. We describe these phases below.

\begin{algorithm}[ht]
 \caption{Hierarchical Annotated Skeleton Planner}
 \label{alg:fullalg}
 \begin{algorithmic}[1]
 \renewcommand{\algorithmicrequire}{\textbf{Input:}}
 \renewcommand{\algorithmicensure}{\textbf{Output:}}
 \REQUIRE Environment $env$, Start $s$, Goal $g$
 \STATE{$ws \leftarrow {\tt BuildWorkspaceSkeleton}(env)$}
 \STATE{$aws \leftarrow {\tt AnnotateSkeletonEdges}(ws)$}
 \STATE{$rdmp \leftarrow {\tt BuildRoadmap}(aws)$}
 \WHILE{$True$}
 \STATE{$pathSet \leftarrow {\tt BuildPathSet}(rdmp, s, g, pathSet)$}
 \IF{$pathFound$}
 \RETURN {$pathSet$}
 \ENDIF
 \WHILE{$\lnot pathSet.empty()$}
 \STATE{$nextBestPath \leftarrow{pathSet.{\tt pop}()}$}
 \STATE{$pathFound \leftarrow {\tt FixPath}(nextBestPath, policy)$}
 \IF{$pathFound$}
 \RETURN {$nextBestPath$}
 \ENDIF
 \STATE{$ {\tt UpdatePathSet}(rdmp, s, g, pathSet)$}
 \ENDWHILE
 \ENDWHILE
 \end{algorithmic}
 \end{algorithm}

\textit{Building the initial roadmap}:
In Line 3 the roadmap is initialized by sampling configurations at each skeleton vertex that satisfies some user-defined acceptance criteria. In the case of a clearance-annotated skeleton, the criterion could be having enough clearance for the robot. We then attempt to connect the samples at each of the skeleton vertices. This step is  the same as the initialization step from \drprm~shown in Figure \ref{fig:drprm_init}.
Each of the connected components $C$ corresponding to a skeleton vertex $v$ is called a local connected component, and we retain pairs $(C, v)$ for future reference. 

For each edge $e$ of the workspace skeleton with endpoints $u$ and $v$, if $e$'s weight satisfies acceptance criteria and if local connected components $C_u$ and $C_v$ were generated at  $u$ and $v$, a roadmap edge is drawn to connect the pair $(C_u, C_v)$ without checking for its validity. Such edges are colored in light blue in Figure \ref{fig:lazy_draw_edges}. At the end of this step, we have a partially validated roadmap with local connected components that are fully validated, and connecting them together are roadmap edges that are guided by the validity of the workspace skeleton.

The next steps are done iteratively until a valid path is found or the allocated computing resources are exhausted (Lines 4-17).

\textit{Building the Path Set}:
First, the start and goal configurations are added to the roadmap. In Line 5 of Algorithm \ref{alg:fullalg}, the algorithm builds a set of paths from the start to goal configurations.

Algorithm \ref{alg:buildpathset} shows how the path set is built. When a potential path is identified, its edges are checked for validity. If a valid path is found, it is returned alone, which triggers the algorithm to exit with success (Lines 6-9 of Algorithm \ref{alg:buildpathset}). Note that additional policies can be added to the "valid path" check (such as path length or quality requirements) as desired. 
If on the other hand, the path contains some invalid edges, as shown in Figure \ref{fig:delete_invalid_edge_in_path}, it is added to the path set by recording its valid and invalid edges as well as its lower bound cost. The lower bound cost is equivalent to the cost that the path would have if all its edges were assumed valid.

 \begin{algorithm}[ht]
 \caption{BuildPathSet}
 \label{alg:buildpathset}
 \begin{algorithmic}[1]
 \renewcommand{\algorithmicrequire}{\textbf{Input:}}
 \renewcommand{\algorithmicensure}{\textbf{Output:}}
 \REQUIRE Roadmap $rdmp$, Start $s$, Goal $g$, 
 Number of desired paths before fixing $maxPaths$,
 Factor of maximum cost limit $\epsilon$,
 Where to save paths $pathSet$
  \STATE{$paths \leftarrow {\tt QueryRoadmap}(rdmp)$}
  \STATE{$maxCost \leftarrow (\epsilon \times paths.last.cost)$} 
  \STATE{$p\_cost = 0$}
  \WHILE{$pathSet.size() < maxPaths ~\AND~p\_cost < maxCost$}
  \STATE{$nextBestPath \leftarrow paths.{\tt pop}()$}
  \IF{${\tt isValid}(nextBestPath)$}
  \STATE{$pathSet \leftarrow nextBestPath$}
  \RETURN $true$
  \ENDIF
  \STATE{$p\_cost = nextBestPath.cost$}
  \STATE{$pathSet.{\tt Add}(nextBestPath)$}
  \ENDWHILE
  \RETURN false
 \end{algorithmic}
 \end{algorithm}

\textit{Fixing the path set}:
Once the path set is built, the planner iterates through the partially-invalid paths in order of least invalid and attempts to fix them one by one (Lines 9 to 14 of Algorithm \ref{alg:fullalg}) using the $FixPath$ method.

In Algorithm \ref{alg:fixpath}, the path's invalid edges are fixed in order of importance (Line 1). The default ordering policy is edge length:  longer edges have higher priority because they have a lower chance of being fixed. This makes them good indicators of whether a path is not fixable.

Once the order is determined, the planner attempts to fix each invalid edge (Lines 2-7). The local components from each end of the skeleton edge are expanded towards each other and reconnected with a shorter lazily drawn edge. The steps to fix an edge are  illustrated by Figures \ref{fig:expand_regions} and \ref{fig:lazy_draw_new_edge}.

If an invalid edge is not valid after an attempt to fix it, the edge is labeled as ``unfixable." The planner updates the path set (Line 15 of Algorithm \ref{alg:fullalg}) to remove  all paths containing the unfixed edge using Algorithm~\ref{alg:UpdatePathSet}. This prevents the planner from trying to fix an unfixable edge multiple times. 
 
 \begin{algorithm}[h]
 \caption{FixPath}
 \label{alg:fixpath}
 \begin{algorithmic}[1]
 \renewcommand{\algorithmicrequire}{\textbf{Input:}}
 \renewcommand{\algorithmicensure}{\textbf{Output:}}
 \REQUIRE Roadmap $rdmp$, Start $s$, Goal $g$, 
 path to fix $path$, policy for fixing path $policy$, where to store unfixable edges $unfixableEdges$
 \STATE{$queue \leftarrow {\tt OrderEdgesToFix}(policy)$} \COMMENT{Queue edges in ascending order}
 \FOR{$edge$ in $queue$}
 \STATE{$isEdgeFixed \leftarrow {\tt FixEdge}(edge, validityCriteria)$}
 \IF{$!isEdgeFixed$}
 \STATE{$unfixableEdges.{\tt Add}(edge)$}
 \ENDIF
 \ENDFOR
 \RETURN $unfixableEdges.{\tt isEmpty}()$
 \end{algorithmic}
 \end{algorithm}

\begin{algorithm}[h]
\caption{UpdatePathSet}
\label{alg:UpdatePathSet}
\begin{algorithmic}[1]
\renewcommand{\algorithmicrequire}{\textbf{Input:}}
 \renewcommand{\algorithmicensure}{\textbf{Output:}}
 \REQUIRE Roadmap $rdmp$, Start $s$, Goal $g$, set of paths $pathSet$, list of unfixable edges (edges which weren't able to get fixed in FixPath) $unfixableEdges$.
 \ENSURE Updates all paths' invalid edge set.
 \FOR{$path$ in $pathSet$}
    \FOR{$edge$ in $unfixableEdges$}
        \IF{$edge$ in $path$}
            \STATE{$pathSet.remove(path)$}
        \ENDIF
    \ENDFOR
 \ENDFOR
\end{algorithmic}
\end{algorithm}

\section{Experimental Validation}

\begin{figure*}[!htb]
\centering
\begin{subfigure}{.3\textwidth}
  \centering
  \includegraphics[width=0.7\textwidth]{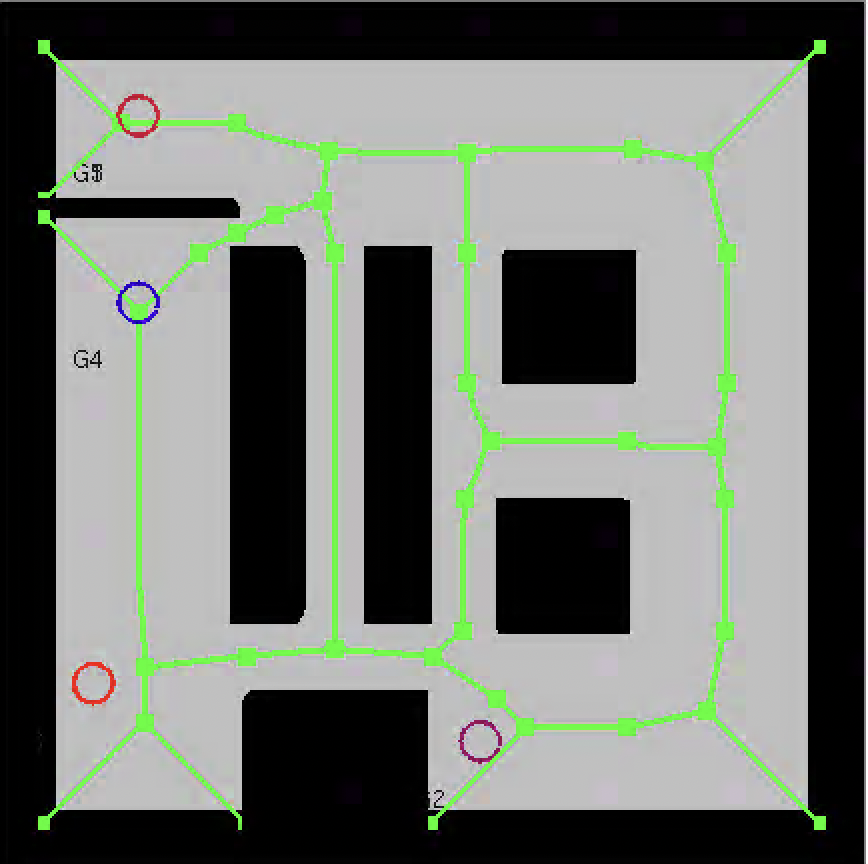}
  \caption{}
  \label{fig:create-env}
\end{subfigure}
 \centering
\begin{subfigure}{.3\textwidth}
  \centering
  \includegraphics[width=1\textwidth]{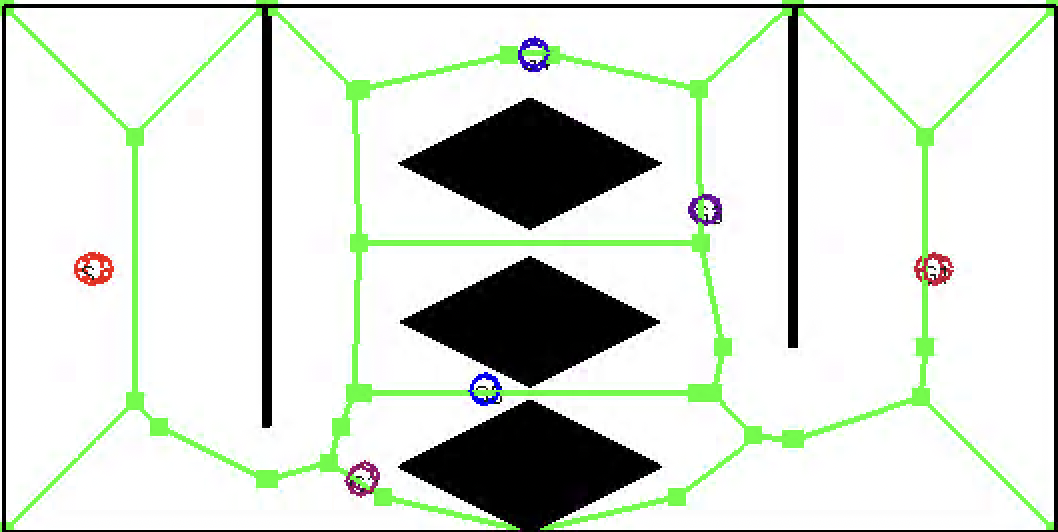}
  \caption{}
  \label{fig:rhombus-env}
\end{subfigure}
\centering
\begin{subfigure}{.3\textwidth}
  \centering
  \includegraphics[width=1\textwidth]{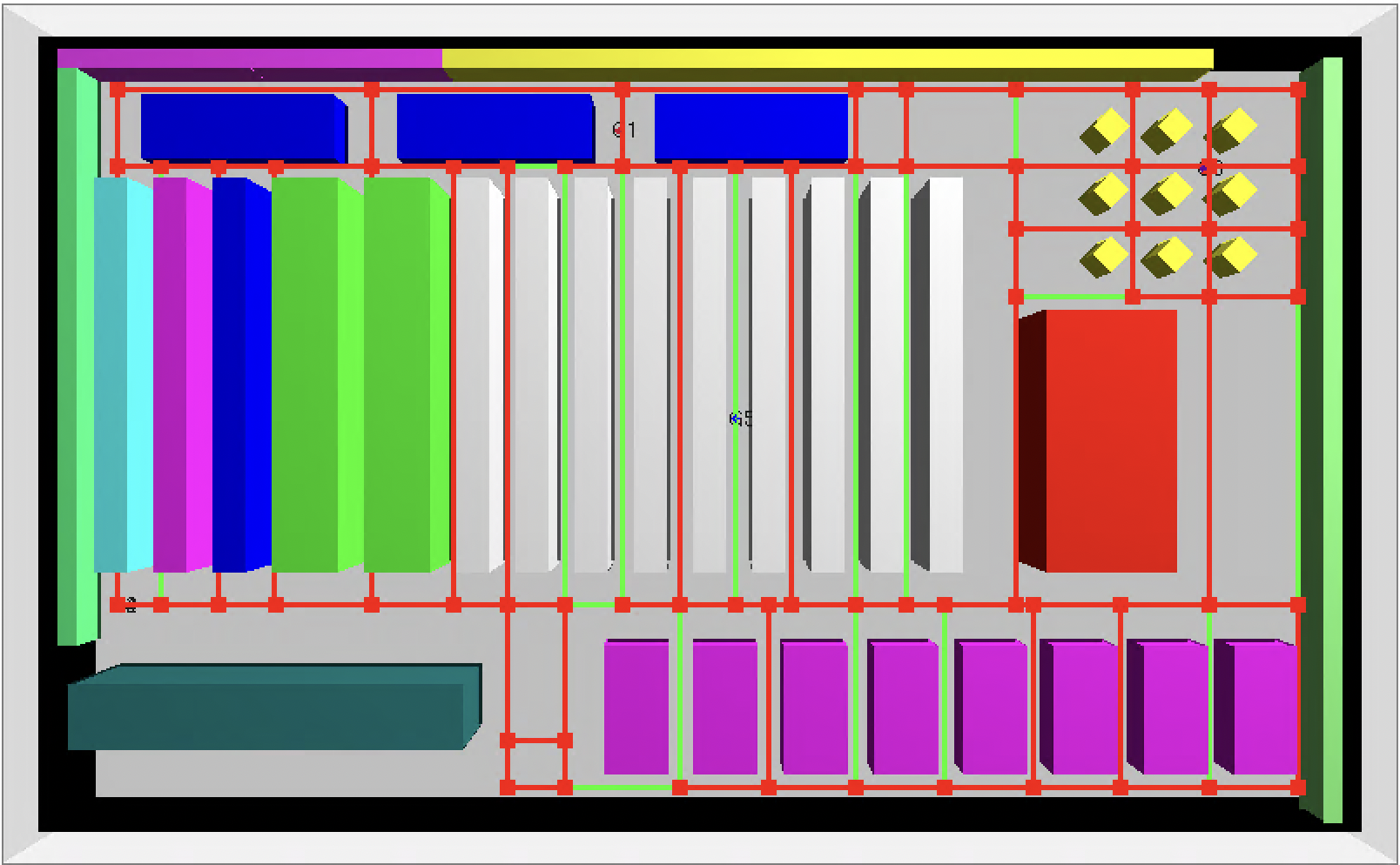}
  \caption{}
  \label{fig:store-env}
\end{subfigure}
\caption{Testing environments. (a) Create, (b) Rhombus, and (c) Store. The medial axis workspace skeleton is shown in green and the traffic-annotated skeleton is color-coded in the Store environment.}
 \label{fig:environments}
\end{figure*}

We evaluate the performance and solution quality from hierarchically guiding path finding with workspace skeleton guidance.
The algorithm is compared to five Probabilistic Roadmaps variants with \prm \cite{kslo-prpp-96} for baseline. Specifically, we compare \hasap~to its predecessor, the skeleton-guided \drprm \cite{suda-tgrcdrs-20}, and to \maprm \cite{lta-gfsmafs-03} which finds maximum clearance paths. In addition, since \hasap~performs lazy evaluations, it is compared to the lazy planners \lazyprm \cite{bk-ppulp-00}, and \customlazyprm, as baselines for postponing validation to the graph search (query) phase. The \customlazyprm~method uses partial validation during roadmap construction.

Three problems were selected to validate the algorithm's performance  as shown in Figure \ref{fig:environments}. The first two were selected to test the ability of the different PRM variants to return the desired path, while the third one tests the scalability of these algorithms.
\begin{itemize}
\item \textbf{iCreate} (Figure \ref{fig:create-env}): A 3 DOF nonholonomic iRobot Create is tested in simulation. In addition to speed, safe navigation is important in this environment with narrow and wide path options. We set the desired path preference to maximum clearance in this environment to study the ability of \hasap~to return the desired solutions in reasonable time.
\item \textbf{Rhombus} (Figure \ref{fig:rhombus-env}): Three rhombuses are placed at different distances from each other to provide pathways with different clearance for a point robot.
\item \textbf{Store} (Figure \ref{fig:store-env}): This is a grocery store environment with a 3 DOF nonholomic robot. The aisles in the store environment have different lengths, which highlights the bottleneck of constraining $\cspace~$ exploration on the intermediates of the workspace skeleton edges. In addition, the environment has fifty obstacles to test the scalability of the skeleton-guided algorithms with respect to the collision detection cost.
\end{itemize}

\subsection{Experiment Setup}
In each environment, three queries were defined to study the performance of different PRM variants in multi-query problems.
Each algorithm was given 1000 sampling attempts to build the initial roadmap, and then runs until it solves all queries.
All algorithms made two samples per iteration to prevent unnecessarily dense roadmaps and used eight nearest neighbors to connect each sample.
Here, we report the time required to build the initial roadmap, the time to solve each query and its path length as a proxy for path cost. In addition, we report the total number of collision detection calls as well as the size of the final roadmap.

The experiments were executed on a desktop computer with an Intel Core i7-3770 CPU at 3.4 GHz, 16 GB of RAM, running CentOS 7  and the GNU g++ compiler version 4.8.5. All methods were implemented in our C++ motion planning library developed in the Parasol Lab at the University of Illinois at Urbana-Champaign.
Roadmap validation was done with the PQP-SOLID collision detection method \cite{lglm-pqp-99}. During roadmap construction, \customlazyprm~was set up to use a partially validating collision detection method, RAPID \cite{glm-rapid-96}. RAPID only checks for the intersection of object surfaces and doesn't check if an object is completely contained inside the obstacle.
Skeletonization for the dynamic region methods was performed with the Medial Axis skeleton \cite{b-tends-67} implemented in the CGAL library \cite{fgkss-dccga-00}. These models are constructed once and read in with the environment by the planner. The time to build the models was considered pre-processing and not included in the results.

\subsection{Analysis}
\label{sec:results_analysis}

\begin{figure*}[!htb]
\centering
\begin{subfigure}{.3\textwidth}
  \centering
  \includegraphics[width=1\textwidth]{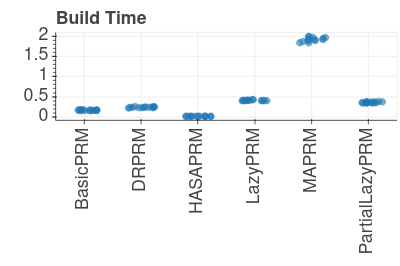}
  \caption{Create}
  \label{fig:create-build}
\end{subfigure}
 \centering
\begin{subfigure}{.3\textwidth}
  \centering
  \includegraphics[width=1\textwidth]{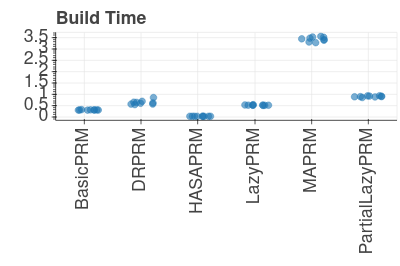}
  \caption{Rhombus}
  \label{fig:rhombus-build}
\end{subfigure}
\centering
\begin{subfigure}{.3\textwidth}
  \centering
  \includegraphics[width=.85\textwidth]{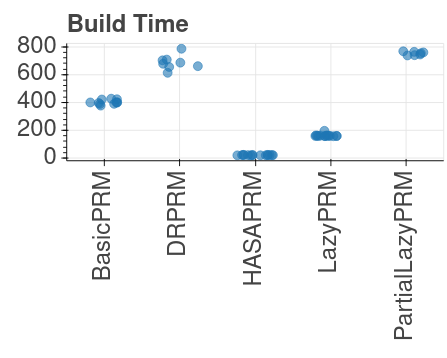}
  \caption{Store}
  \label{fig:store-build}
\end{subfigure}
\caption{Run time in seconds.}
 \label{fig:runtime}
\end{figure*}

\begin{figure*}[!htb]
\centering
\begin{subfigure}{.49\textwidth}
  \centering
  \includegraphics[width=1\textwidth]{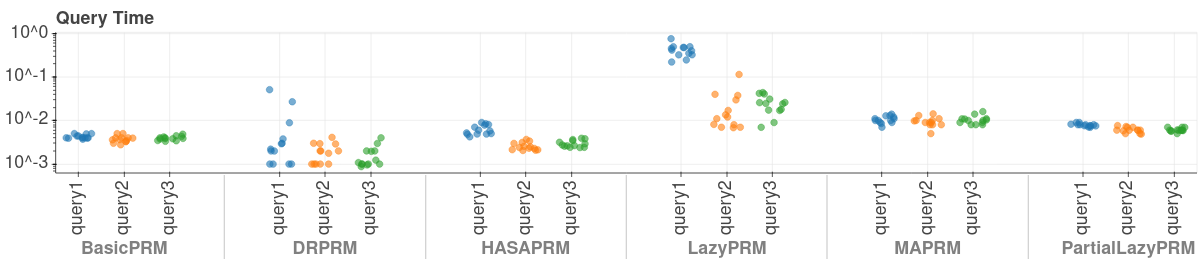}
  \caption{Create - Query Time}
  \label{fig:create-query}
\end{subfigure}
\centering
\begin{subfigure}{.49\textwidth}
  \centering
  \includegraphics[width=1\textwidth]{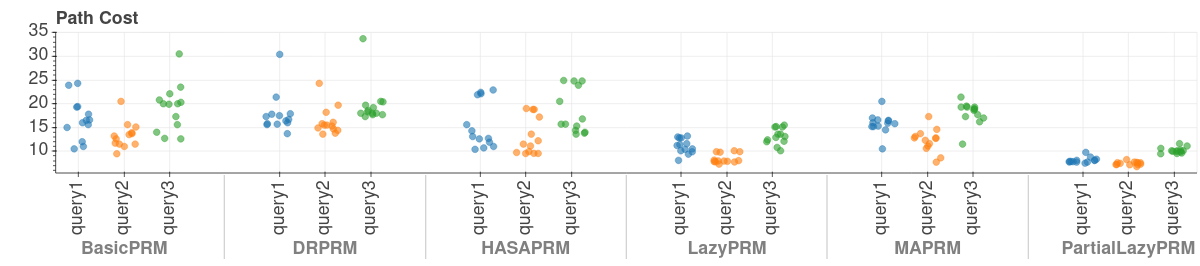}
  \caption{Create - Path Cost}
  \label{fig:create-cost}
\end{subfigure}
~
  \centering
\begin{subfigure}{.49\textwidth}
  \centering
  \includegraphics[width=1\textwidth]{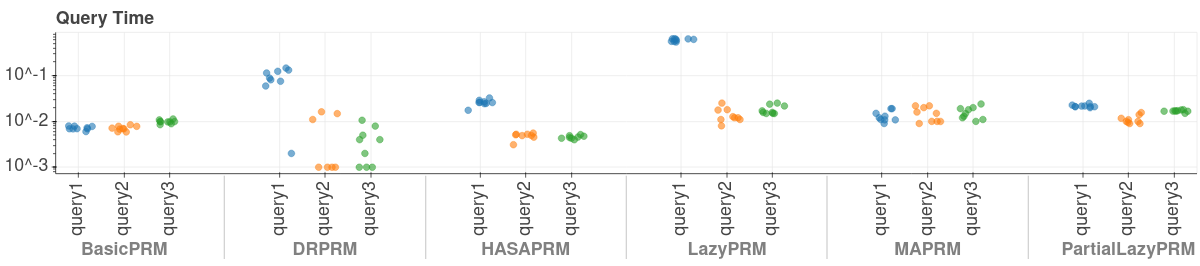}
  \caption{Rhombus - Query Time}
  \label{fig:rhombus-query}
\end{subfigure}
\centering
\begin{subfigure}{.49\textwidth}
  \centering
  \includegraphics[width=1\textwidth]{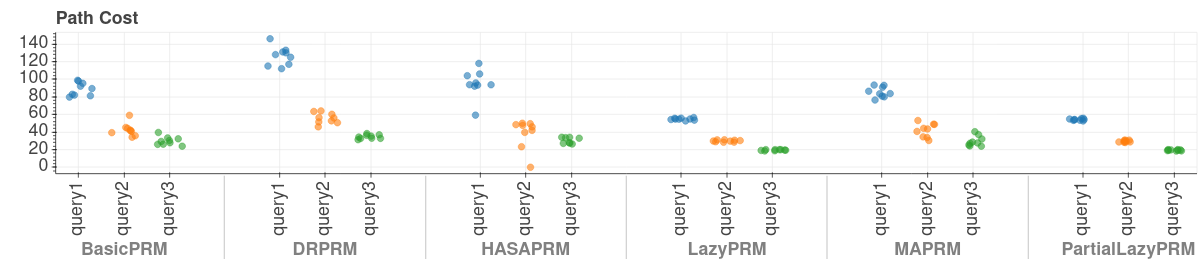}
  \caption{Rhombus - Path Cost}
  \label{fig:rhombus-cost}
\end{subfigure}
~
  \centering
\begin{subfigure}{.49\textwidth}
  \centering
  \includegraphics[width=1\textwidth]{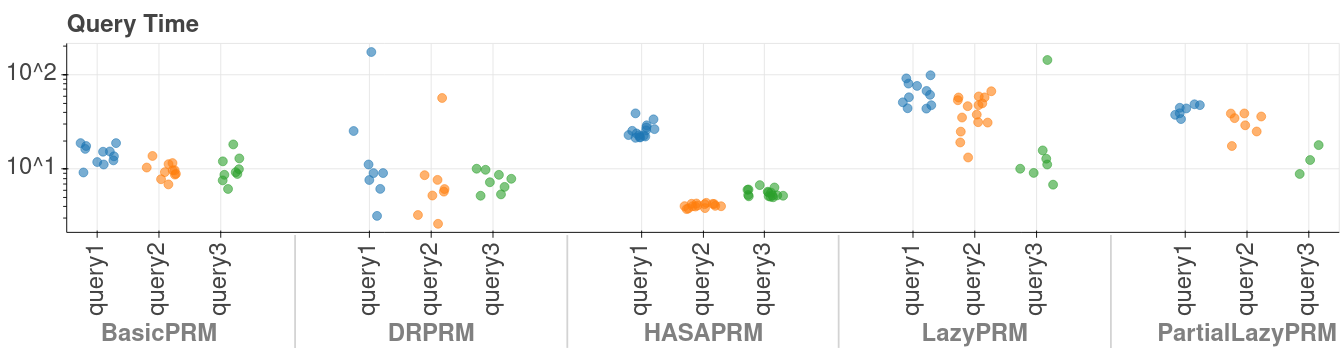}
  \caption{Store - Query Time}
  \label{fig:store-query}
\end{subfigure}
\centering
\begin{subfigure}{.49\textwidth}
  \centering
  \includegraphics[width=1\textwidth]{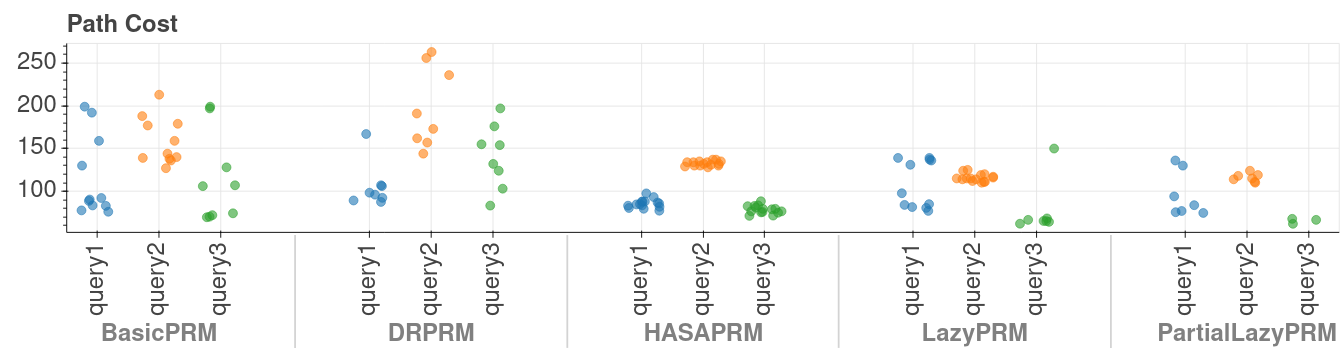}
  \caption{Store - Path Cost}
  \label{fig:store-cost}
\end{subfigure}
\caption{Path query time in seconds and path cost.}
 \label{fig:query-cost-results}
\end{figure*}

As shown in Table~\ref{tab1}, \hasap~ has considerably lower roadmap construction time and cost in the three environments. In addition, Table \ref{tab1} shows that \hasap~solves all queries using the smallest roadmap. \hasap's number of collision detection calls is second to \lazyprm's, which reflects the fact that in addition to validating the path, \hasap~validates its local connected components around skeleton vertices. \hasap~returns high-clearance paths that are close to the ones returned by \maprm, thanks to the lesser number of collision detection needed to build a skeleton-guided roadmap.

\begin{table}[!htbp]
\caption{Roadmap construction results}
\begin{center}
\begin{tabular}{|c|c|c|c|c|}
\hline
\textbf{Environment} & \textbf{Planner} & \textbf{CD calls} & \textbf{Nodes} & \textbf{Edges} \\
\hline
  Create             & Basic-PRM        & 5,667             & 599            & 1,335 \\
  \cline{2-5}
                     & DR-PRM           &  23,556           &  255           &   1,230 \\
  \cline{2-5}
                     & HASP            &  2,523            & \textbf{68}    & \textbf{140} \\
  \cline{2-5}
                     & Lazy-PRM         &  \textbf{585}     &  998         & 15,635 \\
  \cline{2-5}
                     & MAPRM            & 18,031            & 539            & 1,215 \\
  \cline{2-5}
                     & Partial-PRM   &  49,737           &  672           & 9,093 \\
\hline
  Rhombus            &BasicPRM                  &   18,020                &817                &1798        \\\cline{2-5}
                     & DR-PRM & 101,678 & 223 & 1034 \\\cline{2-5}
                     & HASP &  11,517 & \textbf{69}& \textbf{138}\\\cline{2-5}
                     & Lazy-PRM & \textbf{6,886} & 885 & 12,099 \\\cline{2-5}
                     & MAPRM & 70,467 & 892 & 1949\\\cline{2-5}
                     & Partial-PRM & 220,689 & 817 & 12,289\\
  \hline
   Store             & Basic-PRM        & 1,707,796             & 602            & 1,505 \\
  \cline{2-5}
                     & DR-PRM           &  4,706,249           &  3722           &   770 \\
  \cline{2-5}
                     & HASP            & \textbf{1,230,324}            & \textbf{141}    & \textbf{360} \\

  \cline{2-5}
                     & Lazy-PRM         &  5,396,817     &  860         & 10,526 \\
  \cline{2-5}
                     & Partial-PRM   &  5,096,883           &  946           & 6,126 \\                     
\hline
\end{tabular}
\label{tab1}
\end{center}
\end{table}

In the Create environment (Figure \ref{fig:create-query}), \hasap~returns paths in comparable time to the fully validating algorithms like \prm~and \drprm. In addition, the path cost is lower than that of most of the paths returned by \drprm. 
In the Rhombus environment (Figure \ref{fig:rhombus-query}), \hasap~returns paths in less time than the other lazy algorithms and with lowest path cost. Compared to the other methods, the query results from \hasap~have the lowest variance.
In the Store environment (Figure \ref{fig:store-query}), \maprm~could not solve any query within the limit of 1000 attempts and 10 minutes of query time.

\subsubsection{Run time and query time} In all three scenarios in Figure \ref{fig:runtime}, \hasap's planning time is the lowest. Figure \ref{fig:store-build} shows that as the environment gets more cluttered, \hasap's planning time improves by orders of magnitude to that of the other planners. With all the planners in Figure \ref{fig:query-cost-results}, the query time generally improves as more queries get solved, because the roadmap is progressively improved by the resampling done to solve each query. Figure \ref{fig:create-query} and \ref{fig:rhombus-query} show that \hasap's query time is comparable to that of \prm. In the Rhombus and Store environments, \lazyprm~takes longer to solve the first query, because the original roadmap generated in the planning phase mostly occupied the obstacle space in the narrow passages of the environment.  

\subsubsection{Collision Detection Cost} Table \ref{tab1} shows that \hasap~produces sparse roadmaps to solve all the queries. This is correlated with the low number of collision detection calls. In fact, in the Store environment, \hasap~uses close to five times less collision detection calls than the unguided \lazyprm. This is because \hasap's roadmap does not need as much fixing during the query phase to find a valid path. 

\subsubsection{Scalability} The strength of the hierarchical approach is showcased in the Store environment (Figure \ref{fig:store-build}) where \hasap~maintains low planning time as the number of obstacles is increased. In addition, Table \ref{tab1} shows that \hasap~validates its roadmap with less collision detection than \lazyprm.
Given the high number of obstacles in the environment, the planners that skip collision detection usually fare better than their counterparts. However, \lazyprm~suffers in the query time because its roadmap mostly lies in the obstacle space. Moreover, the performance of \drprm~is degraded in this environment because of its reliance on the skeleton and the high cost of building the roadmap. In fact, \drprm~did not build a fully connected roadmap in the 1000 attempts allocated to all the strategies, which resulted in more variability in query time and path cost.

\subsubsection{Dependence on the skeleton}
Planning time and query time in the Store environment show the significance of relying on the skeleton during planning. The long aisles in this environment contain long skeleton edges that constrain \drprm~to merge its local connected components in more steps by following the intermediates of the skeleton edges. The ability of \hasap~to merge the local components faster and fix them as needed, proves to be better suited for this kind of environment.

\section{Discussion}
The results presented in Section \ref{sec:results_analysis} highlight the main strengths of \hasap.
We note that by skipping the localized exploration of regions mapped by skeleton edges, the strategy constructs a sparse roadmap of the $\cspace$ in record time. Although the resulting roadmap is not fully validated, the query time and path cost results
show that skeleton guidance increases the chances of having a roadmap that mostly lies in $\cfree$. 
These results also show that the skeleton, when annotated with properties relevant to the motion planning problem, guides the planner to easily find a desirable path.

Although the presented algorithm was tested in static settings, it could be potentially extended to dynamic settings. The challenges to address for real-time navigation include how to deal with new obstacles in the roadmap structure and in the annotations. Obstacle motions that impact areas not yet validated would not cause any problem, but the validated areas would need to be updated. 
Also, dynamic environment properties can be marked by the annotations with an efficient updating strategy to avoid stalling. We are currently investigating these motion planning problems.

\section{Conclusion}
We present a hierarchical skeleton-guided planning algorithm that initiates local connected components in topological regions of the workspace and connects them using a lazy valid approach and skeleton edge guidance.
The method hierarchically queries the roadmap to find easy solutions that are mapped by the workspace skeleton first, fixing the partially valid paths second and searching for the more difficult paths last if needed.
Experiments show that \hasap~is faster than the other PRM variants to construct a roadmap and that it requires the least dense roadmap to find paths with low path cost. In addition, \hasap~is shown to scale better than the other variants of probabilistic roadmap algorithms, in an environment of fifty obstacles. \par
This work will be extended to guide multi-robot systems by annotating the skeleton with information relevant to each robot like the traffic patterns of other moving agents in the environment. In addition, this work will be extended to incorporate dynamic environment changes to allow the planner to adapt to changes in the workspace by following the guidance of a dynamically annotated skeleton.

\bibliographystyle{styles/IEEEtran}
\bibliography{robotics,biochemistry,geom}
\end{document}